\documentclass[runningheads]{llncs}
\usepackage{graphicx}

\usepackage{tikz}
\usepackage{comment}
\usepackage{amsmath,amssymb} 
\usepackage{color}
\usepackage{multirow}
\usepackage{algorithm}
\usepackage{algpseudocode}
\usepackage{cite}
\usepackage[pagebackref,breaklinks,colorlinks]{hyperref}
\usepackage[capitalize]{cleveref}
\crefname{section}{Sec.}{Secs.}
\Crefname{section}{Section}{Sections}
\Crefname{table}{Table}{Tables}
\crefname{table}{Tab.}{Tabs.}
\usepackage{appendix}

\begin{document}
\pagestyle{headings}
\mainmatter
\def\ECCVSubNumber{100}  

\title{Towards Hard-Positive Query Mining for DETR-based Human-Object Interaction Detection} 
\titlerunning{Towards Hard-Positive Query Mining for DETR-based HOI Detection}

%

\author{Xubin Zhong\inst{1} \and
Changxing Ding\inst{1,2}\thanks{Corresponding author. } \and Zijian Li\inst{1} \and
Shaoli Huang\inst{3}}
%
%
\institute{South China University of Technology , Guangzhou, China \and Pazhou Lab, Guangzhou, China \and
Tencent AI-Lab, Shenzhen, China\\
\email{eexubin@mail.scut.edu.cn, chxding@scut.edu.cn, eezijianli@mail.scut.edu.cn, shaolihuang@tencent.com}}

\maketitle
\vspace{-4mm}

\begin{abstract}
Human-Object Interaction (HOI) detection is a core task for high-level image understanding. Recently, Detection Transformer (DETR)-based HOI detectors have become popular due to their superior performance and efficient structure. However, these approaches typically adopt fixed HOI queries for all testing images, which is vulnerable to the location change of objects in one specific image. Accordingly, in this paper, we propose to enhance DETR’s robustness by mining hard-positive queries, which are forced to make correct predictions using partial visual cues. First, we explicitly compose hard-positive queries according to the ground-truth (GT) position of labeled human-object pairs for each training image. Specifically, we shift the GT bounding boxes of each labeled human-object pair so that the shifted boxes cover only a certain portion of the GT ones. We encode the coordinates of the shifted boxes for each labeled human-object pair into an HOI query. Second, we implicitly construct another set of hard-positive queries by masking the top scores in cross-attention maps of the decoder layers. The masked attention maps then only cover partial important cues for HOI predictions.
Finally, an alternate strategy is proposed that efficiently combines both types of hard queries. In each iteration, both DETR’s learnable queries and one selected type of hard-positive queries are adopted for loss computation. Experimental results show that our proposed approach can be widely applied to existing DETR-based HOI detectors. Moreover, we consistently achieve state-of-the-art performance on three benchmarks: HICO-DET, V-COCO, and HOI-A. Code is available at \texttt{\url{https://github.com/MuchHair/HQM}}.

\end{abstract}

\section{Introduction}
\label{sec:1}
Human-Object Interaction (HOI) detection is a fundamental task for human-centric scene understanding \cite{chao2018learning,ji2020action, lin2022hl,lin2022ru,wang2021batch}.  It aims to infer a set of HOI triplets $<human, interaction, object>$ from a given image \cite{gupta2015visual,chao2018learning}. In other words, it involves  identifying not only the categories and locations of objects in an individual image, but also the interactions between each human–object pair. Recently, Detection Transformer (DETR)-based methods \cite{tamura2021qpic,kim2021hotr,zou2021end,zhang2021mining,qu2022distillation} have become popular in the field of HOI detection due to their superior performance and efficient structure.
These methods typically adopt a set of learnable queries, each of which employs the cross-attention mechanism \cite{vaswani2017attention} to aggregate image-wide context information in order to predict potential HOI triplets at specific locations.

\begin{figure}[t]
\centering
\includegraphics[width=0.88\linewidth]{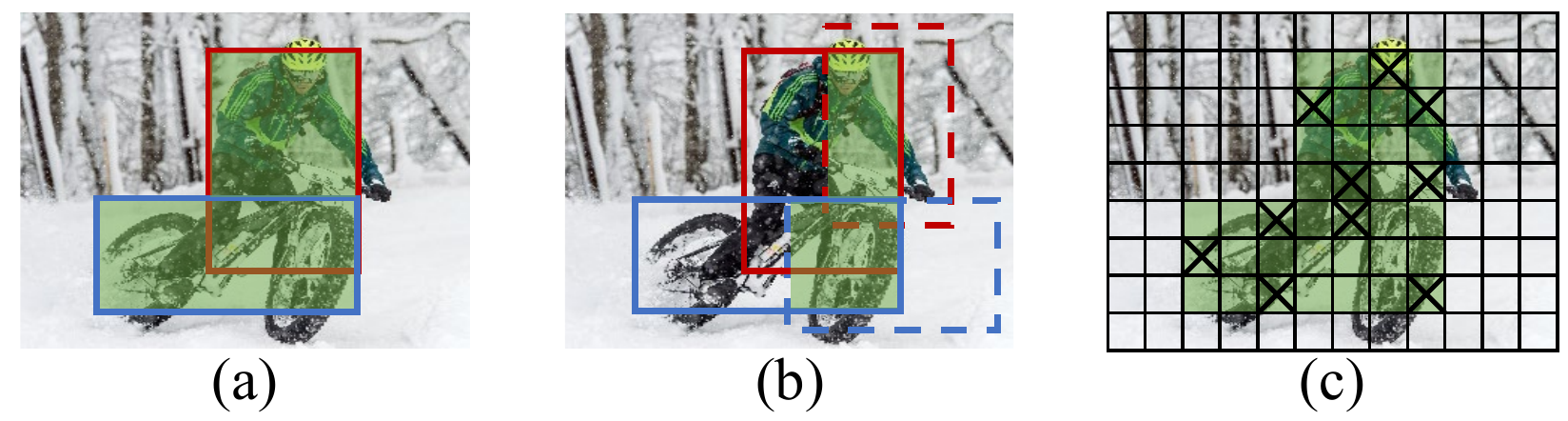}
\caption{Illustration of the hard-positive queries. (a) The green area represents important visual cues for HOI prediction of one human-object pair. (b) The dashed boxes are produced via Ground-truth Bounding-box Shifting (GBS), which only cover part of important image area and are then encoded into hard-positive queries. (c) Part of important visual cues are removed via Attention Map Masking (AMM), which increases the prediction difficulty of one positive query to infer HOI triplets. Best viewed in color.}
\vspace{-5mm}
\label{Figure:1}
\end{figure}

However, the learnable queries are usually weight-fixed after training \cite{qu2022distillation,tamura2021qpic}. Since each query targets a specific location \cite{carion2020end,kim2021hotr}, DETR-based methods are typically sensitive to changes in the object locations in testing images. Recent works improve DETR's robustness through the use of adaptive queries. For example, CDN \cite{zhang2021mining} performs human-object pair detection before interaction classification occurs, generating adaptive interaction queries based on the output of the object detection part. However, its queries for object detection remain fixed. Moreover, two other object detection works \cite{ConditionDetr,liu2022dab} opt to update each object query according to the output embedding of each decoder layer. An object query is typically formulated as a single reference point of one potential object. Notably, this strategy may not be easy to apply in the context of HOI detection, since the interaction area for one human-object pair is usually more complex to formulate \cite{liao2020ppdm,kim2020uniondet,zhong2021glance}.  Therefore, current DETR-based methods still suffer from suffer from poor-quality queries.

In this paper, we enhance the robustness of DETR-based HOI detection methods from a novel perspective, namely that of Hard-positive Query Mining (HQM).  In our approach, the robustness refers to the ability of DETR model to correctly predict HOI instances even using poor-quality queries with limited visual cues (or say inaccurate position). Accordingly, a hard-positive query refers to a query that corresponds to one labeled human-object pair, but is restricted to employ limited visual cues to make correct HOI predictions.
First, as illustrated in Fig. \ref{Figure:1}(b), we explicitly generate such queries via Ground-truth Bounding-box Shifting (GBS). In more detail, we shift the two ground-truth (GT) bounding boxes in one labeled human-object pair so that each shifted box covers only a certain portion of its GT box. We then encode the coordinates of the two shifted boxes into an HOI query. Accordingly, the resultant query contains only rough location information about the pair. This strategy models the extreme conditions caused by variations in object location for fixed HOI queries.

Second, as shown in Fig. \ref{Figure:1}(c), we increase the prediction difficulty of positive queries by means of Attention Map Masking (AMM). The positive queries in AMM are those DETR’s learnable queries matched with ground-truth according to bipartite matching \cite{kuhn1955hungarian}. In more detail, for each positive query, a proportion of the top scores in the cross-attention maps are masked.  In this way, the positive query employs only part of the visual cues for prediction purposes. Specially, as our goal is to enhance the robustness of learnable queries, we select
the masked elements according to the value of their counterparts in the learnable queries. Queries generated via GBS and AMM are less vulnerable to overfitting and capable of producing valuable gradients for DETR-based models. Finally, the robustness of DETR-based models is enhanced for the test images.

During each iteration of the training stage, the DETR’s learnable queries and our hard-positive queries are utilized sequentially for prediction with shared model parameters and loss functions. To promote efficiency, GBS and AMM are alternately selected in each iteration. This Alternate Joint Learning (AJL) strategy is more efficient and achieves better performance than other joint learning strategies. Moreover, during inference, both GBT and AMM are removed; therefore, our method does not increase the complexity of DETR-based models in the testing stage.

To the best of our knowledge, HQM is the first approach that promotes the robustness of DETR-based models from the perspective of hard example mining. Moreover, HQM is plug-and-play and can be readily applied to many DETR-based HOI detection methods. Exhaustive experiments are conducted on three HOI benchmarks, namely HICO-DET \cite{chao2018learning}, V-COCO \cite{gupta2015visual}, and HOI-A \cite{liao2020ppdm}. Experimental results show that HQM not only achieves superior performance, but also significantly accelerates the training convergence speed.

\section{Related Works}
\label{sec:related_work}

\noindent \textbf {Human-Object Interaction Detection.}
Based on the model architecture adopted, existing HOI detection approaches can be divided into two categories: Convolutional Neural Networks (CNN)-based methods  \cite{gupta2019no,Gao2018iCAN,liao2020ppdm} and transformer-based methods
 \cite{tamura2021qpic,kim2021hotr,zou2021end,zhang2021mining,chen2021reformulating}.

CNN-based methods can be further categorized into two-stage approaches \cite{ulutan2020vsgnet,Gao2018iCAN,wang2019deep,zhong2021polysemy,li2019transferable,gupta2019no} and one-stage approaches \cite{liao2020ppdm,kim2020uniondet,zhong2021glance}.
In general terms, two-stage approaches first adopt a pre-trained object detector \cite{ren2015faster} to generate human and object proposals, after which they feed the features of human-object pairs into verb classifiers for interaction prediction.
Various types of features can be utilized to improve interaction classification, including human pose \cite{li2019transferable,gupta2019no}, human-object spatial information \cite{zhou2019relation,wan2019pose,li2020detailed}, and language features \cite{ulutan2020vsgnet,Gao2018iCAN,wang2019deep}.
Although two-stage methods are flexible to include diverse features, they are usually time-consuming due to the cascaded steps.
By contrast, one-stage methods are usually more efficient because they perform object detection and interaction prediction in parallel \cite{liao2020ppdm,kim2020uniondet,zhong2021glance}.
These methods typically depend on predefined interaction areas for interaction prediction. For example, UnionDet \cite{kim2020uniondet} used the union box of a human-object pair as interaction area, while PPDM \cite{liao2020ppdm} employed a single interaction point to represent interaction area. Recently, GGNet \cite{zhong2021glance} utilized a set of dynamic points to cover larger interaction areas.  However, the above predefined interaction areas may not fully explore the image-wide context information.\\
\indent Recently, the transformer architecture has become popular for HOI detection. Most such methods are DETR-based \cite{tamura2021qpic,kim2021hotr,zou2021end,zhang2021mining,chen2021reformulating,yuan2022detecting,li2021improving}. These methods can be further divided into two categories: methods that employ one set of learnable queries for both object detection and interaction classification \cite{tamura2021qpic,zou2021end,yuan2022detecting,li2021improving}, and methods that utilize separate sets of queries for object detection and interaction prediction  \cite{kim2021hotr,chen2021reformulating,zhang2021mining}. The above methods have achieved superior performance through their utilization of image-wide context information for HOI prediction. However, due to their use of weight-fixed queries, their performance is usually sensitive to location change of humans or objects.\\
\noindent \textbf {DETR-based Object Detection.} The DETR model realizes end-to-end object detection by formulating the task as a set prediction problem \cite{carion2020end}. However, due to its use of weight-fixed and semantically  obscure HOI queries queries, it suffers from slow training convergence \cite{liu2022dab,ConditionDetr,ConverDETR,DynamicDETR,DeformableDETR}. To solve this problem, recent works have largely adopted one of two main strategies. The first of these is to impose spatial priors on attention maps in the decoder layers to reduce semantic ambiguity. For example, Dynamic DETR \cite{DynamicDETR} estimates a Region of Interest (ROI) based on the embedding of each decoder layer, then constrains the cross-attention operation in the next decoder layer within the ROI region. The second strategy involves updating queries according to the output decoder embeddings from each decoder layer \cite{ConditionDetr,liu2022dab}. Each query in these works is typically formulated through a single reference point of the object instance. However, it may not be straightforward to make similar formulations in the context of HOI detection. This is because HOI detection is a more challenging task that involves not only the detection of a single object, but also detection of the human instance and interaction category. \\
\indent In this paper, we enhance the robustness of DETR-based models from a novel perspective, namely that of hard-positive query mining. Compared with existing approaches, our method is easier to implement and does not increase model complexity during inference. In the experimentation section, we further demonstrate that our approach achieves better performance than existing methods.\\
\noindent \textbf {Hard Example Mining.} HQM can be regarded as a hard example mining (HEM) approach to transformer-based HOI detection. HEM has demonstrated its effectiveness in improving the inference accuracy of CNN-based object detection models \cite{wang2017fast,shrivastava2016training}. However, this strategy has rarely been explored in HOI detection. Recently, Zhong et al. \cite{zhong2021glance} devised a hard negative attentive loss to overcome the problem of  class imbalance between positive and negative samples for keypoint-based HOI detection models \cite{liao2020ppdm,wang2020learning}. In comparison, HQM modifies DETR’s model architecture in the training stage and resolves the problem caused by its use of poor-quality queries.
\section{Method}

HQM is plug-and-play and can be applied to many existing DETR-based HOI detection models.  \cite{tamura2021qpic,kim2021hotr,zou2021end,zhang2021mining,chen2021reformulating,yuan2022detecting,li2021improving}. In this section, we take the representative work QPIC \cite{tamura2021qpic} as an example. The overall framework of  QPIC \cite{tamura2021qpic} equipped with HQM is shown in Fig.  \ref{Figure:overview}. In the following, we first give a brief review of QPIC, which is followed by descriptions of two novel hard-positive
query mining methods, e.g., GBS  (Section \ref{HPQ}) and AMM (Section \ref{AMM}). Finally, we introduce an alternate joint learning strategy to  apply GBS and AMM (Section \ref{ATS}) efficiently.

\subsection{Overview of Our Method}
\begin{figure*}[t]
	\centering
		\includegraphics[width=1.0\textwidth]{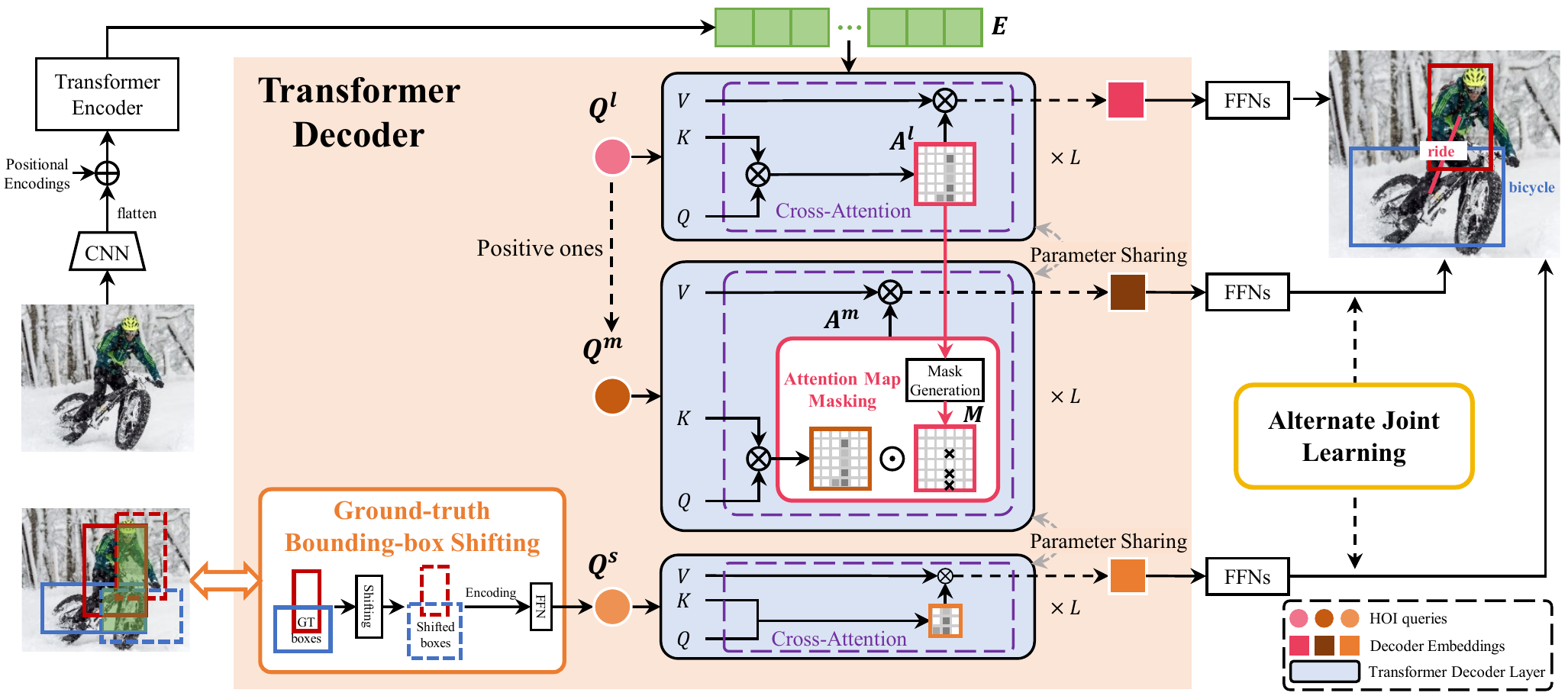}
	\caption{Overview of HQM in the training stage based on QPIC \cite{tamura2021qpic}.
	In the interest of simplicity, only one learnable query $\textbf{Q}^{l}$ and two hard-positive queries are illustrated. The hard-positive query $\textbf{Q}^{s}$ is produced by GBS that encodes the coordinates of shifted bounding boxes of one human-object pair into a query. Another positive query $\textbf{Q}^{m}$ is selected from the learnable queries according to bipartite matching with the ground-truth. The cross-attention maps of $\textbf{Q}^{m}$ are partially masked to increase prediction difficulty. The two types of hard-positive queries are alternately selected in each iteration, and the chosen type of queries are utilized together with the learnble queries for loss computation. $\otimes$ and $\odot$ denote matrix multiplication and Hadamard product, respectively. In the inference stage, HQM is removed and therefore brings no extra computational cost. Best viewed in color.}
	\label{Figure:overview}
	\vspace{-3mm}
\end{figure*}
\noindent \textbf{Revisit to QPIC. }
As shown in Fig. \ref{Figure:overview}, QPIC is constructed by a CNN-based backbone, a transformer encoder, a transformer decoder, and feed-forward networks (FFNs).
Each input image \textbf{I} $\in \mathbb{R}^{H_0 \times W_0 \times 3}$ is first fed into the CNN backbone and the transformer encoder to extract flattened visual features $\textbf{E} \in  \mathbb{R}^{(H \times W) \times D}$. QPIC then performs cross-attention between learnable queries $\textbf{Q}^{l} \in \mathbb{R}^{N_{q}\times D}$ and $\textbf{E}$ in the transformer decoder. $N_{q}$ is the number of learnable queries fixed after training. $H \times W$ and $D$ denote the number of the image patches and feature dimension for each patch, respectively. Besides, the transformer decoder is typically composed of multiple stacked layers. For clarity, we only present the cross-attention operation in one decoder layer. The output embedding $\textbf{C}_{i} \in \mathbb{R}^{N_{q} \times D }$ of the $i$-th decoder layer can be formulated as follows:
\begin{align}
    \textbf{C}_{i} &= \operatorname{Concat}([\textbf{A}^{l}_{h} \textbf{E} \textbf{W}_{h}^V]^T_{h=1}), \\
    \textbf{A}^{l}_{h} &= \operatorname{Softmax}(Att_{h}(\textbf{Q}^{l}, \textbf{C}_{i-1},\textbf{E})),
    \label{attention map}
\end{align}
where $T$ is the number of cross-attention heads. $\textbf{A}^{l}_{h} \in \mathbb{R}^{N_{q} \times (H\times W)} $ is the normalized cross-attention map for the $h$-th head. $\textbf{W}_{h}^V$ is a linear projection matrix. $Att_{h}(\cdot)$ is a function for similarity computation. Finally, as illustrated in Fig. \ref{Figure:overview}, each output decoder embedding is sent to detection heads based on FFNs to obtain object class scores, interaction category scores, and position of the human and object instances.

\noindent \textbf{Hard-Positive Query Mining.} Most DETR-based HOI detection methods, e.g., QPIC, adopt poor-quality queries after training. As analyzed in Section \ref{sec:1}, performance of weight-fixed queries is sensitive to the location  change of human and object instances in testing images.
In the following, we propose to promote the robustness of DETR-based models via hard-positive query mining. One hard-positive query refers to a query that corresponds to one labeled human-object pair, but is restricted to employ limited visual cues to infer correct HOI triplets (shown in Fig. \ref{Figure:1}).

As illustrated in Fig. \ref{Figure:overview}, two strategies are introduced to produce hard-positive queries, i.e., GBS and AMM. Their generated queries are denoted as $\textbf{Q}^{s}  \in \mathbb{R}^{N_{g}\times D} $ and $\textbf{Q}^{m} \in \mathbb{R}^{N_{g}\times D} $, respectively. $N_{g}$ is the number of labeled human-object pairs in one training image. Similar to $\textbf{Q}^{l}$, $\textbf{Q}^{s}$ and $\textbf{Q}^{m}$ are sent to the transformer decoder and their output decoder embeddings are forced to infer correct HOI triplets. $\textbf{Q}^{l}$, $\textbf{Q}^{s}$ and $\textbf{Q}^{m}$ share all model layers and loss functions.

\subsection{Ground-truth Bounding-box Shifting} \label{HPQ}
Previous works have shown that each query attends to specific locations in one image \cite{carion2020end,kim2021hotr}. To enhance DETR's robustness against the spatial variance of human-object pairs, we here explicitly compose hard-positive queries $\textbf{Q}^{s}$ according to the GT position of labeled human-object pairs for each training image.
As shown in Fig. \ref{Figure:1}(b), we shift the bounding boxes of labeled human-object pairs such that the shifted boxes only contain partial visual cues for HOI prediction.

Specifically, we encode a hard-positive query $\textbf{q}^{s} \in \textbf{Q}^{s}$  for one labeled human-object pair as follows:

\begin{equation}
\textbf{q}^{s} = L_{n}(F_{p}(Shift(\textbf{p}^{s} ))), \\
\label{eq1}
\end{equation}
where
\begin{equation}
\textbf{p}^{s} = [x_h, y_h, w_h, h_h, x_o, y_o, w_o, h_o, x_h-x_o, y_h-y_o, w_hh_h, w_oh_o]^{T}. \\
\label{eq2}
\end{equation}
The first eight elements in $\textbf{p}^{s}$ are the center coordinates, width, and height of one GT human-object pair, respectively. [$x_h-x_o$, $y_h-y_o$] denotes the relative position between the two boxes, respectively; while the last two elements are the areas of the two boxes. $Shift(\cdot)$ represents the shifting operation to GT bounding boxes (as illustrated in Fig. \ref{Figure:overview}). $F_{p}(\cdot)$ is an FFN with two layers whose dimensions are both $D$. It projects $\textbf{p}^{s}$ to another $D$-dimensional space. $L_{n}(\cdot)$ is a $tanh$ normalization function, which  ensures the amplitudes of elements in $\textbf{q}^{s}$ and the positional embeddings for $\textbf{E}$  are consistent \cite{qu2022distillation}.

Compared with one concurrent work DN-DETR \cite{li2022dn}, GBS focuses on hard-positive query mining. To ensure the query is both positive and hard, we control the Intersection-over-Unions (IoUs) between each shifted box and its ground truth. We adopt low IoUs ranging from 0.4 to 0.6 in our experiment and find that GBS significantly improves the inference performance of DETR-based models.

\begin{algorithm}[t]
\renewcommand{\algorithmicrequire}{\textbf{Input: }}
\caption{Attention Map Masking for Each Attention Head} \label{pseudocode}
\begin{algorithmic}[1]
\State \algorithmicrequire attention maps $\textbf{A}^{m}$, $\textbf{A}^{l} \in \mathbb{R}^{H \times W}$ for a hard-positive query, ${K}$, $\gamma$
\State Get the indices $I_{K}$ of the top-${K}$ elements in $\textbf{A}^{l}$
\State
Initialize a random binary mask $\textbf{M} \in \mathbb{R}^{H \times W}$: $\textbf{M}_{i, j} \sim Bernoulli(\gamma)$

\For {$\textbf{M}_{i, j} \in \textbf{M}$}
    \If {$(i, j) \notin I_{K}$}
        \State {${\textbf{M}}_{i, j}$ = 1 }
    \EndIf
\EndFor
\State \textbf{Output: } Masked attention map $\textbf{A}^{m} = \textbf{A}^{m} \odot \textbf{M}$
\end{algorithmic}
\end{algorithm}
\subsection{Attention Map Masking}
\label{AMM}
One popular way to enhance model robustness is Dropout \cite{srivastava2014dropout}. However, applying Dropout directly to features or attention maps of $\textbf{Q}^{l}$ may cause interference to bipartite matching \cite{kuhn1955hungarian}, since the feature quality of the query is artificially degraded. To solve this problem, we implicitly construct another set of hard-positive queries $\textbf{Q}^{m}$ via AMM after the bipartite matching of $\textbf{Q}^{l}$. Queries in $\textbf{Q}^{m}$ are copied from the positive queries in $\textbf{Q}^{l}$ according to results by bipartite matching.
As shown in  Fig. \ref{Figure:overview}, to increase the prediction difficulty of $\textbf{Q}^{m}$, some
elements in the cross-attention maps for each query in $\textbf{Q}^{m}$ are masked. In this way, each query in $\textbf{Q}^{m}$ is forced to capture more visual cues from the non-masked regions.\\
\indent Detailed operations of AMM are presented in Algorithm \ref{pseudocode}. For clarity, only one hard-positive query $\textbf{q}^{m} \in \textbf{Q}^{m}$ is taken as an example, whose attention maps are denoted as $\textbf{A}^{m}_{h}$  (1 $\leq h \leq T$). For simplicity, we drop the subscripts of both $\textbf{A}^{m}_{h}$ and $\textbf{A}^{l}_{h}$ in the following.\\
\indent AMM has two parameters, i.e., $K$ and $\gamma$. Since our ultimate goal is to enhance the robustness of   $\textbf{Q}^{l}$ rather than $\textbf{Q}^{m}$, we select dropped elements in ${\textbf{A}}^{m}$ according to the value of their counterparts in ${\textbf{A}}^{l}$.  Specifically, we first select the top $K$ elements according to the value in ${\textbf{A}}^{l}$. Then, we randomly mask the selected $K$ elements in ${\textbf{A}}^{m}$ with a ratio of $\gamma$.\\
 \noindent \textbf{Discussion. } AMM is related but different from Dropout and its variants \cite{srivastava2014dropout, ghiasi2018dropblock}. Their main difference lies in the way to select dropped elements. First, AMM drops elements with high values, while Dropout drops elements randomly. Second, AMM requires a reference, i.e., ${\textbf{A}}^{l}$, for dropped element selection in ${\textbf{A}}^{m}$. Dropout requires no reference. In the experimentation section, we show that AMM achieves notably better performance than the naive Dropout.

\subsection{Alternate Joint Learning} \label{ATS}
The two hard-positive query mining methods, i.e., GBS and AMM, can be applied  jointly to generate diverse hard queries.
However, as DETR-based HOI detection methods typically require large number of training epochs to converge, it is inefficient to adopt both methods together in each iteration. We here present the Alternate Joint Learning (AJL) strategy, in which GBS and AMM are applied alternately for each training iteration. Concretely, the learnable queries of DETR and our hard queries are fed into the transformer decoder sequentially. The main reason for this lies in the design of AMM.  The
masked attention scores for hard queries are selected according to those of learnable queries (in Section \ref{AMM}). Therefore, learnable queries should pass the model first to provide attention scores. Compared to applying GBS and AMM together for each iteration, AJL is more efficient and achieves better performance in our experiments.

\noindent \textbf{Overall Loss Function. } We adopt the same loss functions for object detection and interaction prediction as those in QPIC \cite{tamura2021qpic}. The overall loss function in the training phase can be represented as follows:
\begin{equation}
\centering
\begin{aligned}
 \mathcal{L} =  \alpha \mathcal{L} _{l} + \beta \mathcal{L} _{h},
\end{aligned}
\label{eq:all_loss}
\vspace{-2mm}
\end{equation}
where
\begin{equation}
\centering
\begin{aligned}
  \mathcal{L} _{l} = \lambda_{b}  \mathcal{L} _{l_{b}}  +  \lambda_{u}  \mathcal{L} _{l_{u}}  +
  \lambda_{c}  \mathcal{L} _{l_{c}} +   \lambda_{a}  \mathcal{L} _{l_{a}},
\end{aligned}
\label{eq:all_loss2}
\vspace{-2mm}
\end{equation}

\begin{equation}
\centering
\begin{aligned}
  \mathcal{L} _{h} = \lambda_{b}  \mathcal{L} _{h_{b}}  +  \lambda_{u}  \mathcal{L} _{h_{u}}  +
  \lambda_{c}  \mathcal{L} _{h_{c}} +   \lambda_{a}  \mathcal{L} _{h_{a}}.
\end{aligned}
\label{eq:all_loss3}
\end{equation}
$  \mathcal{L}_{l}$  and $ \mathcal{L}_{h}$ denote the loss for learnable queries and hard-positive queries, respectively.
$ \mathcal{L}_{k_{b}}$,
$ \mathcal{L}_{k_{u}}$,
$ \mathcal{L}_{k_{c}}$, and
$ \mathcal{L}_{k_{a}}$ $(k \in \{l, h\})$ denote the L1 loss,
GIOU loss \cite{GIOU} for bounding box regression,
cross-entropy loss for object classification,
and focal loss \cite{lin2017focal} for interaction prediction,
respectively.
These loss functions are realized in the same way as in \cite{tamura2021qpic}.
Moreover, both $\alpha$ and $\beta$ are set as 1 for simplicity;
while $\lambda_{b}$, $\lambda_{u}$, $\lambda_{c}$ and $\lambda_{a}$ are set as 2.5, 1, 1, 1, which are the same as those in \cite{tamura2021qpic}.

\section{Experimental Setup}

\subsection{Datasets and Evaluation Metrics}

\noindent \textbf{HICO-DET. }
HICO-DET \cite{chao2018learning} is the most popular large-scale HOI detection dataset, which provides more than 150,000 annotated instances. It consists of 38,118 and 9,658 images for training and testing, respectively. There are 80 object categories, 117 verb categories, and 600 HOI categories in total.


\noindent \textbf{V-COCO. }
V-COCO \cite{gupta2015visual}  was constructed based on the MS-COCO database \cite{lin2014microsoft}.
The training and validation sets contain 5,400 images in total, while its testing set includes 4,946 images.
It covers 80 object categories, 26 interaction categories, and 234 HOI categories.
The mean average precision of Scenario 1 role (\emph{mAP$_{role}$}) \cite{gupta2015visual} is commonly used for evaluation.

\noindent \textbf{HOI-A. } HOI-A  was recently proposed in \cite{liao2020ppdm}. The images are collected from wild; it is composed of 38,629 images, with 29,842 used for training and 8,787 for testing.  HOI-A contains 11 object categories and 10 interaction categories.

\subsection{Implementation Details}
We adopt ResNet-50  \cite{he2016deep} as our backbone model. Following QPIC  \cite{tamura2021qpic}, we initialize parameters of our models using those of DETR that was  pre-trained on the MS-COCO database as an object detection task. We adopt the AdamW \cite{loshchilov2018decoupled} optimizer and conduct experiments with a batch size of 16 on 8 GPUs.  The initial learning rate is set as 1e-4 and then decayed to 1e-5 after 50 epochs;  the total number of training epochs is 80.
$N_{q}$ and $D$ are set as 100 and 256, respectively. For GBS, the IoUs between the shifted and ground-truth bounding boxes range from 0.4 to 0.6; while for AMM, $K$ and $\gamma$ are set as 100 and 0.4, respectively.


\subsection{Ablation Studies}
We perform ablation studies on HICO-DET, V-COCO, and HOI-A datasets to demonstrate the effectiveness of each proposed component. We adopt QPIC  \cite{tamura2021qpic} as the baseline and all experiments are performed using ResNet-50 as the backbone. Experimental results are tabulated in Table~\ref{tab:tab1}.

\begin{figure*}[t]
\centering
\includegraphics[width=\linewidth]{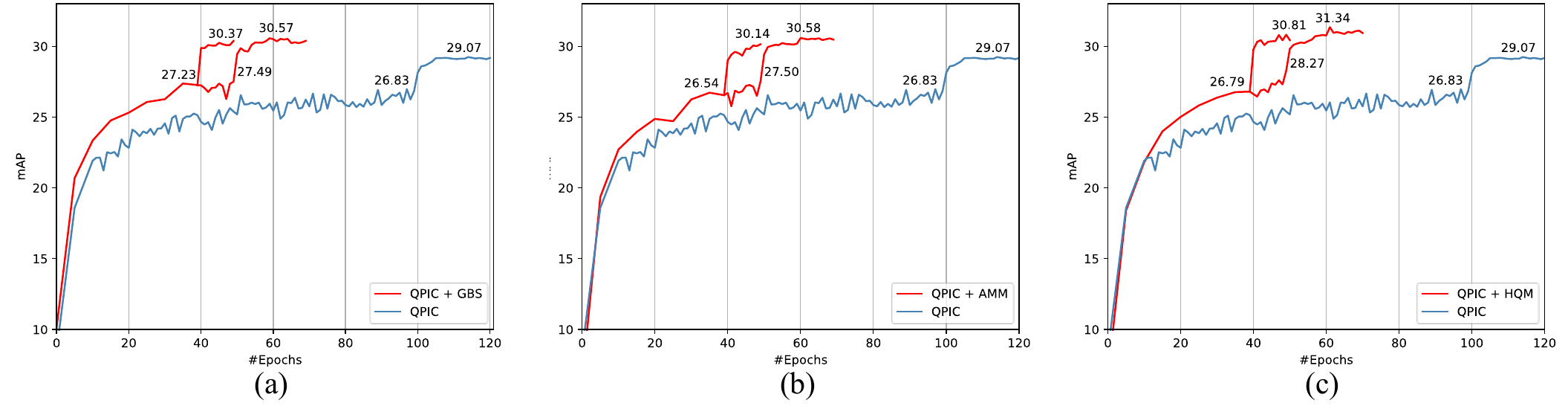}
\vspace{-5mm}
\caption{The mAP and training convergence curves for QPIC  and our method on HICO-DET. Our method significantly improves QPIC in both mAP
accuracy and convergence rate. }
\label{Figure:4}
\end{figure*}

\noindent{\bf{Effectiveness of GBS.}} GBS is devised to explicitly generate hard-positive queries leveraging the bounding box coordinates of labeled human-object pairs. When GBS is incorporated, performance of QPIC is promoted by 1.50\%, 1.39\% and 1.13\% mAP on HICO-DET, V-COCO, and HOI-A datasets, respectively. Moreover, as shown in Fig. \ref{Figure:4}(a), GBS also significantly accelerates the training convergence of QPIC. It justifies the superiority of GBS in improving DETR-based HOI detectors. We further evaluate the optimal values of IoUs and provide experimental results in the supplementary material.

\noindent{\bf{Effectiveness of AMM.}}  AMM is proposed to implicitly construct hard-positive queries using masking to cross-attention maps. As illustrated in Table~\ref{tab:tab1}, the performance of QPIC is notably improved by 1.51\%, 1.48\%, and 1.20\% mAP on HICO-DET, V-COCO, and HOI-A datasets, respectively. Furthermore, as shown in Fig. \ref{Figure:4}(b), AMM also significantly reduces the number of training epochs required on the HICO-DET dataset.
 We also provide a detailed analysis for  $K$ and $\gamma$  values in the supplementary material.

\noindent{\bf{Combination of GBS and AMM.}} We here investigate three possible strategies that combine GBS and AMM for more effective DETR training, namely, Cascaded Joint Learning (CJL), Parallel Joint Learning (PJL) and Alternate Joint Learning (AJL).

\textbf{Cascaded Joint Learning.} In this strategy, we formulate GBS and AMM as two successive steps to produce one single set of hard-positive queries. In more details, we first apply GBS to produce one set of hard-positive queries. Then, we apply AMM to cross-attention maps of queries generated by GBS. As shown in Table~\ref{tab:tab1}, CJL achieves worse performance than the model using GBS or AMM alone. This may be because queries generated by CJL contain rare cues for HOI prediction, thereby introducing difficulties in  optimizing DETR-based models.

\textbf{Parallel Joint Learning.} In this strategy, GBS and AMM are employed to generate one set of hard-positive queries, respectively. Then, both sets of hard-positive queries are employed for HOI prediction. To strike a balance between the loss of learnable queries and hard-positive queries, the loss weight for each type of hard-positive queries is reduced by one-half. Moreover, they are independent, which means there is no interaction between these two types of queries. As shown in Table \ref{tab:tab1}, PJL achieves better performance than the model using GBS or AMM alone. Moreover, it outperforms QPIC by 1.74\%, 1.59\%, and 1.49\% mAP on HICO-DET, V-COCO, and HOI-A datasets, respectively. However, PJL lowers the computational efficiency due to the increased number of hard-positive queries.

\begin{table*}[t]
\centering
\caption{Ablation studies on each key component of HQM. For HICO-DET, the DT mode is adopted for evaluation. }
 \vspace{-2mm}
	\resizebox{0.88\textwidth}{!}{
		\begin{tabular}{c ccccc ccc c c}
			\hline
			& \multicolumn{5}{c}{Components}  &\multicolumn{3}{c}{mAP}  &\multicolumn{1}{c}{\# Epochs} \\
			Method     &GBS  &AMM  &CJL & PJL  &AJL   & HICO-DET   & V-COCO & HOI-A &HICO-DET    \\
			\hline
		     Baseline&-&-&-&-&-   &29.07           &61.80 &74.10 &150 \\
		     \hline
			\multirow{4}{*}{Incremental }
			
			&\checkmark&-&-&-&-    &30.57         & 63.19 &75.23   &80  \\
            &-&\checkmark&-&-&-   &30.58  &63.28 &75.30 &80\\
           &\checkmark&\checkmark&\checkmark&-&-      &30.11      & 63.03  &75.01  &80 \\
           &\checkmark&\checkmark&-&\checkmark&-     &30.81      & 63.39  &75.59  &80 \\
           \hline
            Our Method&\checkmark&\checkmark&-&-&\checkmark  &\textbf{31.34}        &\textbf{63.60} &\textbf{76.13}   &80 \\
			\hline
	   \end{tabular}}
        \label{tab:tab1}
\vspace{-3mm}
\end{table*}

\textbf{Alternate Joint Learning.} In this strategy, GBS and AMM are applied alternately for each training iteration. The learnable queries of DETR and our hard-positive queries are fed into the transformer decoder sequentially, meaning there is no interference between each other.  As tabulated in Table \ref{tab:tab1}, AJL outperforms other joint learning strategies. AJL also has clear advantages in efficiency compared with PJL. Moreover, it significantly promotes the performance of QPIC by 2.27\%, 1.80\%, and 2.03\% mAP on the three datasets,  respectively. The above experimental results justify the effectiveness of AJL.

\noindent\textbf{Application to Other DETR-based Models.} Both GBS and AMM are plug-and-play methods that can be readily applied to other DETR-based HOI detection models, e.g., HOTR \cite{kim2021hotr} and CDN \cite{zhang2021mining}. The main difference between HOTR \cite{kim2021hotr}  and QPIC is that HOTR performs object detection and interaction prediction in parallel branches with independent queries. Here, we mainly apply HQM to its interaction detection branch. As presented in Table \ref{tab:tab2}, HOTR+GBS (AMM) outperform HOTR by 1.21\% (1.27\%) mAP in DT mode for the full HOI categories. When AJL is adopted, the performance of HOTR is considerably improved by  2.23\%, 8.49\% and 0.36\% mAP in DT mode for the full, rare and non-rare HOI categories, respectively. Besides, significant improvements can also be observed by applying our methods to CDN.  Impressively, when incorporated with our method, performance of CDN is  promoted by 1.03\% mAP for the full HOI categories.

\begin{table}[t]
\centering
\begin{minipage}{0.5\textwidth}
    \centering
\caption{Effectiveness of GBS and AMM on HOTR and CDN in the DT Mode of HICO-DET. }
	\resizebox{0.92\linewidth}{!}{
            \begin{tabular}{c ccc ccc}
			\hline
			& \multicolumn{3}{c}{Incremental Components}  &\multicolumn{3}{c}{mAP}  \\
			Baseline     &GBS  &AMM &AJL  &Full& Rare&Non-rare  \\
			\hline
			\multirow{4}{*}{HOTR}
			 &-&-&- &23.46 &16.21 &25.62\\
			&\checkmark&-&- &24.67    &23.29 &25.34   \\
            &-&\checkmark&-&24.73     & 23.52 &25.09 \\
            &\checkmark&\checkmark&\checkmark     & \textbf{25.69} & \textbf{24.70} &  \textbf{25.98}  \\
        		\hline
        		\hline
			\multirow{4}{*}{CDN}
			 &-&-&- &31.44 & 27.39 & 32.64 \\
			&\checkmark&-&-& 32.07 & 27.52 & 33.43   \\
            &-&\checkmark&-& 32.05 & 27.15 & 33.51 \\ 
            &\checkmark&\checkmark&\checkmark & \textbf{32.47} & \textbf{28.15} & \textbf{33.76}   \\
			\hline
    		\end{tabular}
		}
        \label{tab:tab2}
    \end{minipage}
     \hfill
    \vspace{-6mm}
    \begin{minipage}{0.48\textwidth}
    \centering
    \caption{Comparisons with variants of GBS on HICO-DET.}
        \vspace{-2mm}
        \resizebox{0.88\linewidth}{!}{
            \begin{tabular}{l| ccc}
        		\hline
        			&		Full& Rare&Non-rare\\
        		\hline
        		\hline
        		QPIC \cite{tamura2021qpic}	 & 29.07 &21.85 &31.23  \\
        		\hline
        		w/o $Shift(\cdot)$ &29.61 	&22.67  &31.68\\
        		w  Gaussian  noise   & 30.05 & 24.08  & 31.82\\
        		\hline
        		QPIC	+  GBS  &   \textbf{30.57}	& \textbf{24.64}    &\textbf{32.34} \\
        		\hline
    		\end{tabular}
		}
	\label{table:OLT}	
    \end{minipage}
\end{table}

\subsection{Comparisons with Variants of GBS and AMM}
\subsubsection{Comparisons with Variants of GBS.}
We compare the performance of GBS with its two possible variants. Experimental results are tabulated in Table \ref{table:OLT}.

First, `w/o $Shift(\cdot)$' means removing the box-shifting operation in Eq. (\ref{eq2}). This indicates that the ground-truth position of one human-object pair is leveraged for query encoding. Therefore, the obtained queries in this setting are easy-positives rather than hard-positives. It is shown that the performance of this variant is lower than our GBS by 0.96\%, 1.97\% and 0.66\% mAP in DT mode for the full, rare and non-rare HOI categories respectively. This experimental result provides direct evidence for the effectiveness of  hard-positive queries.

\begin{table}[t]
\centering
\begin{minipage}{0.5\textwidth}
    \centering
 \caption{Comparisons with variants of AMM on HICO-DET.}
        \vspace{-1mm}
        \resizebox{0.92\linewidth}{!}{
            \begin{tabular}{l| ccc}
        		\hline
        			&		Full& Rare&Non-rare\\
        		\hline
        		\hline
        	QPIC \cite{tamura2021qpic}  & 29.07 &21.85 &31.23  \\
        		\hline
        	 w/o top-$K$  &30.06  &24.10 &31.84 \\
        	w/o $\textbf{A}^{l}$  &30.11  &24.28 &31.85 \\
        	w/o $\textbf{Q}^{m}$ & 28.75  & 21.97 & 30.78 \\
        	 \hline
        	 QPIC \cite{tamura2021qpic} + AMM &   \textbf{30.58}	&\textbf{25.48} &\textbf{32.10} \\
        		\hline
    		\end{tabular}
		}
		\vspace{2mm}
	\label{table:AMM}
    \end{minipage}
     \hfill
    \vspace{-6mm}
    \begin{minipage}{0.48\textwidth}
    \centering
     \caption{Performance Comparisons on HOI-A. D-based is short for DETR-based.}
            \vspace{-2mm}
        \resizebox{0.8\linewidth}{!}{
            \begin{tabular}{c|c|c|c}
        \hline
        &Methods    & Backbone  & mAP \\
        \hline
        \hline
        \multirow{5}*{\rotatebox{90}{CNN-based}}
        &iCAN \cite{Gao2018iCAN} &ResNet-50 &44.23 \\
        &TIN \cite{li2019transferable} &ResNet-50 &48.64 \\
        &GMVM \cite{GMVM} &ResNet-50 &60.26\\
        &C-HOI \cite{C-HOI} &ResNet-50 &66.04\\
        &PPDM \cite{liao2020ppdm} &Hourglass-104 &71.23\\
        \hline
        \hline
        \multirow{3}*{\rotatebox{90}{D-based}}
        &AS-Net \cite{chen2021reformulating} &ResNet-50 &72.19\\
        &QPIC  \cite{tamura2021qpic} &ResNet-50 &74.10\\
          &QPIC \cite{tamura2021qpic}  + \textbf{HQM} &ResNet-50 &\textbf{76.13}\\
        \hline
        \end{tabular}
		}\label{HOI-A}
    \end{minipage}
\end{table}

Second, `w Gaussian noise' represents that we remove $Shift(\cdot)$ and add Gaussian noise to $\textbf{q}^{s}$ in Eq. \ref{eq1}. This variant provides another strategy to generate hard-positive queries.  Table \ref{table:OLT} shows that GBS outperforms this variant by 0.52\% mAP for the full HOI categories. The main reason is that operations in GBS are more explainable and physically meaningful than adding random Gaussian noise.This experiment justifies the superiority of GBS for producing hard-positive queries. \\
\noindent{\bf{Comparisons with Variants of AMM.}} We here compare the performance of AMM with some possible variants, namely, `w/o top-$K$', `w/o $\textbf{A}^{l}$', and `w/o $\textbf{Q}^{l}$'.  Experimental results are tabulated in Table~\ref{table:AMM}. \\
\indent First, `w/o top-$K$' is a variant that  randomly masks elements rather than the large-value elements in an attention map with the same  $\gamma$ ratio. We can observe that the performance of this variant is lower than AMM by 0.52\% in terms of DT mAP for the full HOI categories. Compared with this variant,  AMM is more challenging since  visual cues are partially removed. Therefore, AMM forces each query to explore more visual cues in unmasked regions, which avoids overfitting. This experiment further demonstrates the necessity of mining hard queries.\\
\indent Second, `w/o $\textbf{A}^{l}$'  means that we select the masked elements according to $\textbf{A}^{m}$ rather than $\textbf{A}^{l}$ in Algorithm \ref{pseudocode}. Compared with AMM, the mAP of this variant drops by 0.47\%, 1.20\%, 0.25\% for the full, rare and non-rare HOI categories. This may be because the learnable queries rather than the hard-positive queries are employed during inference. Therefore, masking according to $\textbf{A}^{l}$ can push the hard-positive queries to explore complementary features to those attended by the learnable ones. In this way, more visual cues can be mined and the inference power of learnable queries can be enhanced during inference.\\
\indent Finally,  `w/o $\textbf{Q}^{m}$' indicates that we apply the same masking operations as AMM to the attention maps of $\textbf{Q}^{l}$ rather than those of $\textbf{Q}^{m}$. In this variant,
$\textbf{Q}^{m}$ are removed and only $\textbf{Q}^{l}$ are adopted as queries. It is shown that the performance of this setting is significantly lower than those of AMM. As analyzed in Section \ref{AMM}, applying dropout directly to attention maps of $\textbf{Q}^{l}$ may degrade the quality of their decoder embeddings, bringing in interference to bipartite matching and therefore causing difficulties in optimizing the entire model.

\subsection{Comparisons with State-of-the-art Methods}
\noindent{\bf{Comparisons on HICO-DET}}. As shown in Table \ref{tab:hico}, our method outperforms all state-of-the-art methods by considerable margins. Impressively, QPIC + HQM outperforms QPIC by 2.27\%, 4.69\%, and 1.55\% in mAP on the full, rare and non-rare HOI categories of the DT mode, respectively. When our method is applied to HOTR and CDN-S with ResNet-50 as the backbone, HOTR (CDN-S) + HQM achieves a 2.23\% (1.03\%) mAP performance gain in DT mode for the full HOI categories over the HOTR (CDN-S) baseline. These experiments justify the effectiveness of HQM in enhancing DETR's robustness. Comparisons under the KO mode are presented in the supplementary material.

\begin{table}[t]
\centering
\begin{minipage}{0.6\textwidth}
    \centering
\caption{Performance comparisons on HICO-DET.}
 \vspace{-3mm}
\resizebox{0.8\textwidth}{!}{
\begin{tabular}{c|c|c|ccc  }
\hline
                &    & & \multicolumn{3}{c}{DT Mode} \\
& Methods & Backbone & Full & Rare & Non-Rare  \\
\hline
\hline
\multirow{7}*{\rotatebox{90}{CNN-based}}
&InteractNet \cite{Gkioxari2017Detecting} & ResNet-50-FPN   &9.94  & 7.16  & 10.77 \\
&UnionDet \cite{kim2020uniondet}& ResNet-50-FPN   &17.58  & 11.72  & 19.33 \\
& PD-Net \cite{zhong2020polysemy}  & ResNet-152  &20.81 &15.90 &22.28\\
& DJ-RN \cite{li2020detailed} & ResNet-50  &21.34 &18.53 &22.18 \\
&PPDM \cite{liao2020ppdm}& Hourglass-104   & 21.73 &13.78 &24.10 \\
& GGNet \cite{zhong2021glance}    & Hourglass-104   &23.47 &16.48 &25.60\\
& VCL \cite{hou2020visual}  & ResNet-101  &23.63 &17.21 &25.55 \\

\hline
\hline
\multirow{9}*{\rotatebox{90}{DETR-based}}
&HOTR \cite{kim2021hotr}  & ResNet-50  &23.46 &16.21 &25.62 \\
&HOI-Trans \cite{zou2021end}   & ResNet-50 & 23.46 & 16.91 & 25.41 \\
 &AS-Net \cite{kim2020uniondet}   & ResNet-50 &28.87  & 24.25  & 30.25 \\
&QPIC  \cite{tamura2021qpic}  & ResNet-50   &29.07 &21.85 &31.23  \\
&ConditionDETR \cite{ConditionDetr}    & ResNet-50 &29.65 &22.64 &31.75  \\
&CDN-S \cite{zhang2021mining}     & ResNet-50  &31.44 & 27.39 & 32.64 \\
 &HOTR \cite{kim2021hotr}  + \textbf{HQM}     & ResNet-50  & \textbf{25.69} & \textbf{24.70} &  \textbf{25.98}  \\
&QPIC \cite{tamura2021qpic} + \textbf{HQM}  & ResNet-50  &\textbf{31.34} &\textbf{26.54} &\textbf{32.78} \\
& CDN-S \cite{zhang2021mining}   + \textbf{HQM}   & ResNet-50 & \textbf{32.47} & \textbf{28.15} & \textbf{33.76}  \\
\hline
\end{tabular}}
\label{tab:hico}
    \end{minipage}
     \hfill
    \vspace{-6mm}
    \begin{minipage}{0.36\textwidth}
    \centering
       \caption{Performance comparisons on V-COCO.}
                \vspace{-2mm}
        \resizebox{0.88\textwidth}{!}{
        \begin{tabular}{c|c|c|c}
        \hline
        &Methods    & Backbone   & $AP_{role}$ \\
        \hline
        \hline
        \multirow{10}*{\rotatebox{90}{CNN-based}}
        &DRG \cite{Gao-ECCV-DRG}  &ResNet-50-FPN &51.0 \\
        &PMFNet \cite{wan2019pose} &ResNet-50-FPN & 52.0 \\
        &PD-Net \cite{zhong2020polysemy} &ResNet-152 &52.6\\
        &ACP \cite{kim2020detecting} &ResNet-152 &52.9\\
        &FCMNet  \cite{liuamplifying} &ResNet-50&53.1\\
        &ConsNet \cite{liu2020consnet}  &ResNet-50-FPN  &53.2\\
        &InteractNet \cite{Gkioxari2017Detecting} &ResNet-50-FPN  &40.0 \\
        &UnionDet \cite{kim2020uniondet} &ResNet-50-FPN &47.5\\
        &IP-Net  \cite{wang2020learning} &Hourglass-104 &51.0\\
       & GGNet \cite{zhong2021glance}    &Hourglass-104 & 54.7 \\
        \hline
        \hline
        \multirow{6}*{\rotatebox{90}{DETR-based}}
        &HOI-Trans \cite{zou2021end} &ResNet-101  &52.9 \\
        &AS-Net \cite{chen2021reformulating} &ResNet-50 &53.9\\
        &HOTR \cite{kim2021hotr} &ResNet-50  &55.2 \\
        &QPIC  \cite{tamura2021qpic} & ResNet-50 &58.8\\
         &CDN-S  \cite{zhang2021mining} & ResNet-50 &61.7\\
          &QPIC \cite{tamura2021qpic}  + \textbf{HQM}  & ResNet-50 &\textbf{63.6} \\

        \hline
        \end{tabular}}
        \label{VCOCO}	
    \end{minipage}
\end{table}

Moreover, we compare the performance of HQM with Conditional DETR \cite{ConditionDetr}. Conditional DETR relieves the weight-fixed query problem via updating queries according to decoder embeddings in each decoder layer. We extend this approach to HOI detection by using an interaction point to represent one potential human-object pair. To facilitate fair comparison, all the other settings are kept the same for HQM and Conditional DETR. Table \ref{tab:hico} shows that HQM achieves better performance. The above experiments justify the superiority of HQM for improving the robustness of DETR-based models in HOI detection.

\noindent{\bf{Comparisons on HOI-A}}. Comparison results on the HOI-A database are summarized in Table  \ref{HOI-A}. The same as results on HICO-DET, our approach outperforms all state-of-the-art methods. In particular, QPIC + HQM significantly outperforms QPIC by 2.03\% in mAP when the same backbone adopted.

\noindent{\bf{Comparisons on V-COCO}}. Comparisons on V-COCO are tabulated in Table \ref{VCOCO}. It is observed that our method still outperforms all other approaches, achieving 63.6\% in terms of $AP_{role}$.
These experiments demonstrate that HQM can consistently improve the robustness of DETR-based models on HICO-DET, VCOCO, and HOI-A datasets.

\subsection{Qualitative Visualization Results}
\begin{figure*}[t]
    \centering
    \includegraphics[width=1.0\textwidth]{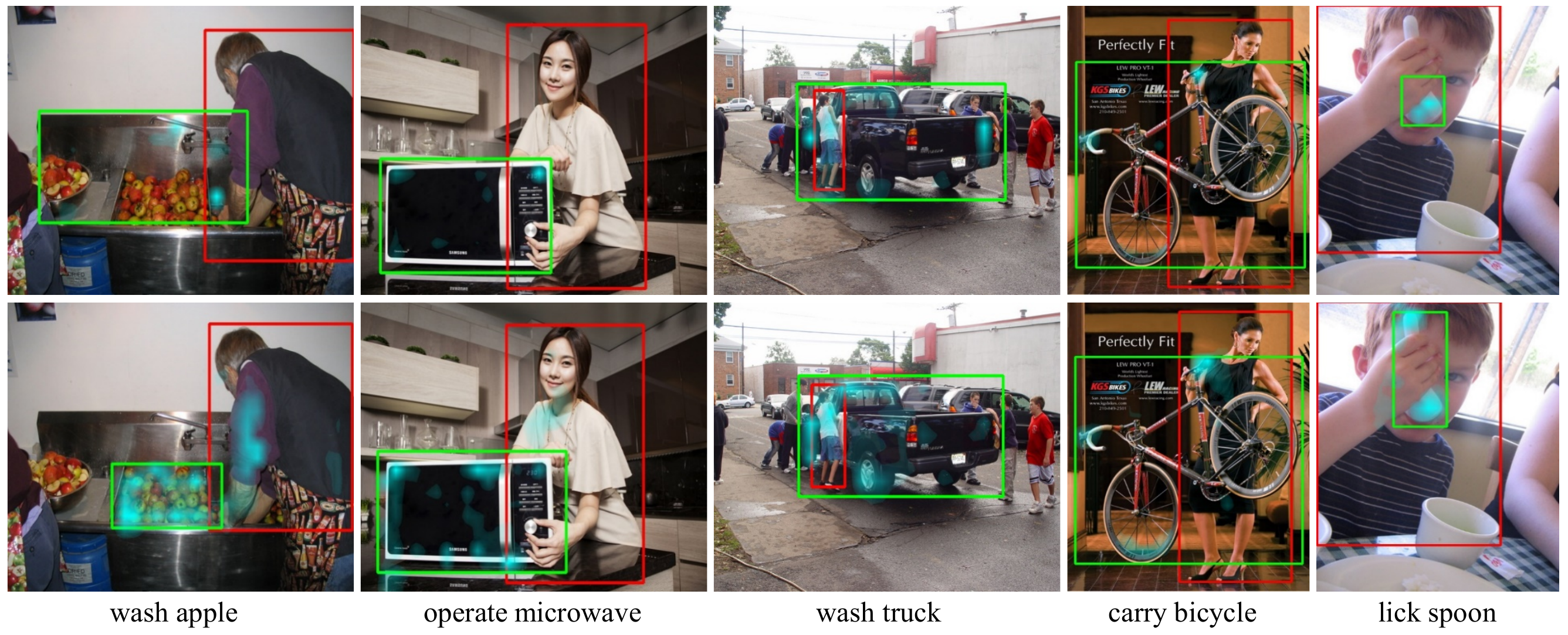}
   \caption{Visualization of HOI detection results and the cross-attention maps in one decoder layer on HICO-DET. Images in the first and second rows represent results for QPIC and QPIC+HQM, respectively. Best viewed in color.}
    \label{Figure:Visualization}
    \vspace{-1mm}
\end{figure*}

As presented in Fig. \ref{Figure:Visualization}, we visualize some HOI detection results and cross-attention maps from  QPIC (the first row) and QPIC + HQM (the second row). It can be observed that QPIC + HQM captures richer visual cues. One main reason for this may be that QPIC + HQM is trained using hard-positive queries. Therefore, QPIC + HQM is forced to mine more visual cues to improve the model prediction accuracy during inference. More qualitative comparisons results are provided in the supplementary material.

\section{Conclusions}
This paper promotes the robustness of existing DETR-based HOI detection models. We creatively propose Hard-positive Queries Mining (HQM) that enhances the robustness of DETR-based models from the perspective of hard example mining. HQM is composed of three key components: Ground-truth Bounding-box Shifting (GBS), Attention Map Masking (AMM), and Alternate Joint Learning (AJL). GBS explicitly encodes hard-positive queries leveraging coordinates of shifted bounding boxes of labeled human-object pairs. At the same time, AMM implicitly constructs hard-positive queries by masking  high-value elements in the cross-attention scores. Finally, AJL is adopted to alternately select one type of the hard positive queries during each iteration for efficiency training. Exhaustive ablation studies on three HOI datasets are performed to demonstrate the effectiveness of each proposed component. Experimental results show that our proposed approach can be widely applied to existing DETR-based HOI detectors. Moreover, we consistently achieve state-of-the-art performance on three benchmarks: HICO-DET, V-COCO, and HOI-A.
\section*{Acknowledgements}
This work was supported by the National Natural Science Foundation of China under Grant 62076101 and 61702193, the Program for Guangdong Introducing Innovative and Entrepreneurial Teams under Grant 2017ZT07X183, Guangdong Basic and Applied Basic Research Foundation under Grant 2022A1515011549, and Guangdong Provincial Key Laboratory of  Human Digital Twin under Grant 2022B1212010004.

\clearpage
%
%
\bibliographystyle{splncs04}

\clearpage

\section*{Supplementary Material}

\begin{appendix}
This supplementary material includes four sections.
Section \ref{Ablation} conducts ablation study on the value of some hyper-parameters in our approach.  Section \ref{app} illustrates the structure and convergence curve of applying our methods to HOTR \cite{kim2021hotr} and CDN-S \cite{zhang2021mining}. Section \ref{sec:hico} presents the complete comparison results.  Section \ref{vis} provides qualitative comparison results for QPIC \cite{tamura2021qpic} and QPIC + HQM.

\section{Ablation Study on Hyper-parameters}
\label{Ablation}
\subsection{Ablation Study on the Value of IoUs in GBS}
\label{GBS}
Experiments are conducted on the HICO-DET database \cite{chao2018learning}. Results are summarized in Table \ref{tab:GBS}.
It is shown that GBS achieves the best performance when the IoUs are randomly sampled within $[$0.4, 0.6$]$.
\begin{table}[!ht]
\centering
\begin{minipage}{0.5\textwidth}
    \centering
\caption{Ablation study on the value of IoUs in GBS  in DT Mode of HICO-DET. }
	\resizebox{0.7\linewidth}{!}{
            \begin{tabular}{l| ccc}
        		\hline
        		IoUs	&		Full& Rare&Non-rare\\
        		\hline
        		\hline
        		$[$0.3, 0.5$]$	 & 30.30&24.63  &32.01  \\
        			$[$0.4, 0.6$]$ &   \textbf{30.57}	& \textbf{24.64}    &\textbf{32.34} \\
        			$[$0.5, 0.7$]$  &30.35 &24.55 &32.08\\
        		\hline
    		\end{tabular}
		}
        \label{tab:GBS}

    \end{minipage}
     \hfill
    \vspace{-6mm}
    \begin{minipage}{0.48\textwidth}
    \centering
    \caption{Ablation study on the value of $K$ and $\gamma$ in DT Mode of HICO-DET.}
        \vspace{-2mm}
        \resizebox{0.7\linewidth}{!}{
            \begin{tabular}{cc| ccc}
        		\hline
        			$K$  &$\gamma$		&Full& Rare&Non-rare\\
        		\hline
        		\hline
        		 100	&0.2  & 30.46 & 25.30 & 32.00 \\
        	 	 100	&0.4  &   \textbf{30.58}	&\textbf{25.48} &\textbf{32.10} \\
                 100    &0.6  & 30.38 & 24.74 & 32.07 \\
        		\hline
        		\hline
                 80  &0.4	 &30.39  &25.34  &31.90   \\
                 100 &0.4	 & \textbf{30.58}	&\textbf{25.48} &\textbf{32.10}  \\
                 120 &0.4	 &30.24 &24.30   &32.00\\
        		\hline
    		\end{tabular}
		}
	\label{table:AMM}	
    \end{minipage}
\end{table}

\subsection{Ablation Study on the Value of $K$ and $\gamma$ in AMM}
\label{AMM}
Experiments are conducted on the HICO-DET database and the results are tabulated in Table \ref{table:AMM}.
We observe that AMM achieves the best performance when  $K$ and $\gamma$ are set as 100 and 0.4, respectively.

\section{Application to HOTR and CDN-S}
\label{app}

HQM is plug-and-play and can be readily applied to many DETR-based HOI detection methods.
We here present more results on applying HQM  to HOTR \cite{kim2021hotr}  and CDN-S \cite{zhang2021mining}.

HOTR is composed of a CNN backbone, a transformer encoder, an instance decoder, an interaction decoder, and interaction detection heads. HOTR performs human-object pair detection and interaction prediction in parallel branches with independent queries. We mainly apply HQM to its interaction detection branch to overcome the weight-fixed query problem.

Compared with HOTR, CDN-S formulates human-object pair detection and interaction prediction as two successive steps. Decoder embeddings produced by the former step are adopted as queries for the latter one. Therefore, queries for the interaction decoder have been adaptive rather than weight-fixed. Accordingly, we apply HQM to CDN’s decoder layers for human-object pair detection, which adopt weight-fixed queries.

\begin{figure*}[t]
\centering
\includegraphics[width=0.88\textwidth]{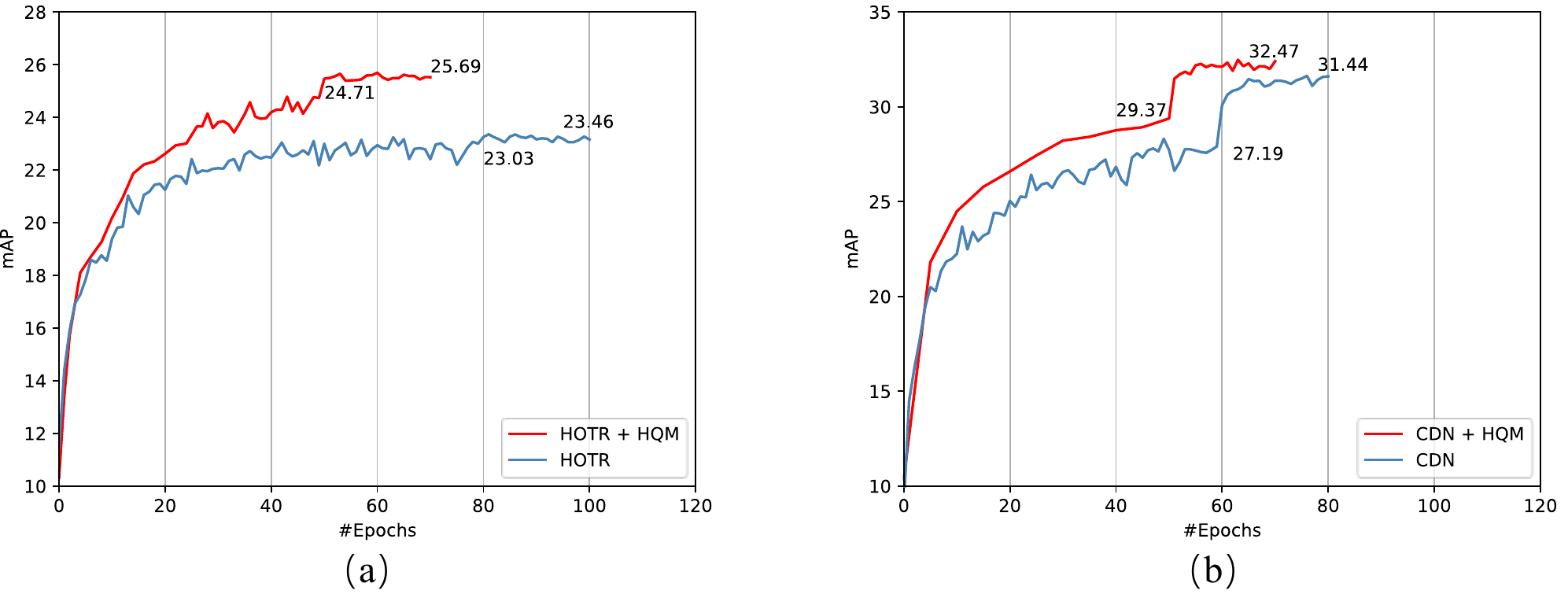}
\vspace{-5mm}
\caption{The mAP and training convergence curves on HICO-DET. (a) Results  for HOTR and HOTR + HQM. (b) Results for CDN-S and CDN-S + HQM. }
\label{Figure:conv}
\end{figure*}

We will release codes to show more details of applying HQM to HOTR and CDN-S. As shown in Fig. \ref{Figure:conv}, HQM not only promotes the mAP accuracy of both models, but also significantly accelerates their training convergence rates.

\begin{figure*}[t]
    \centering
    \includegraphics[width=1.0\textwidth]{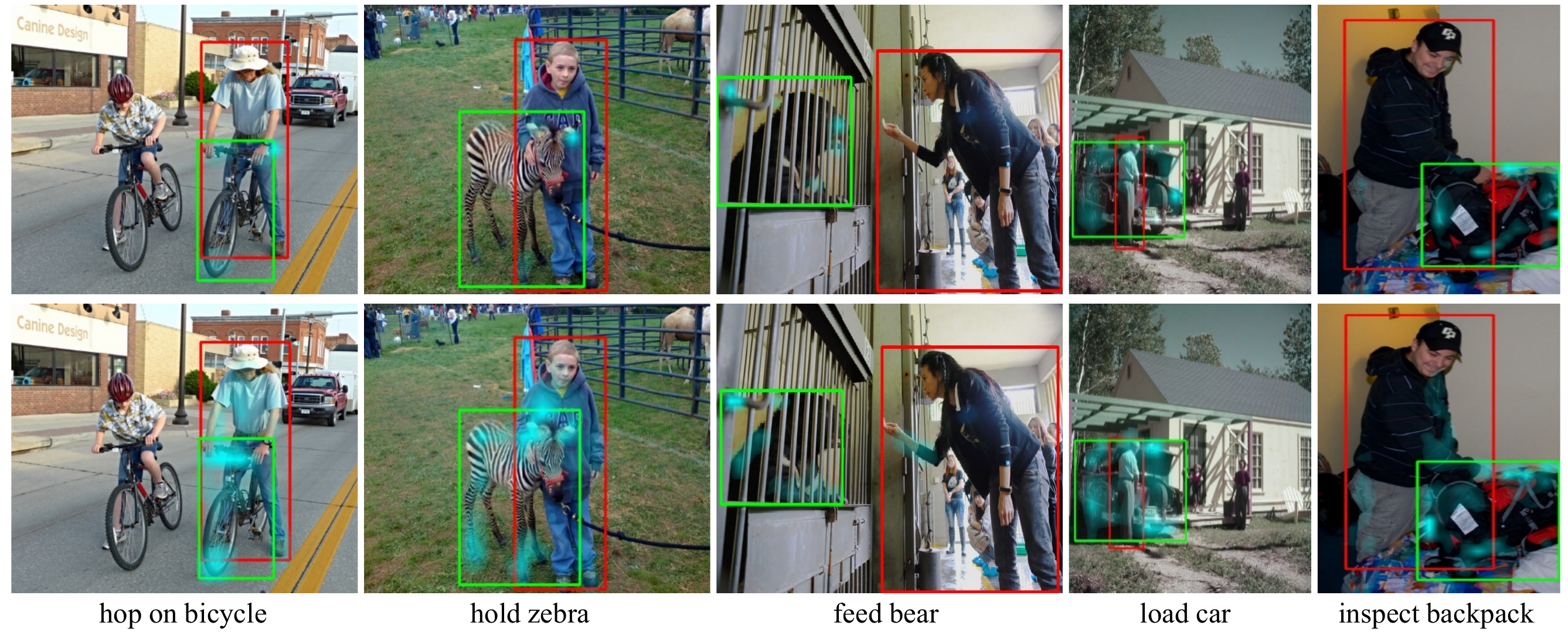}
   \caption{Visualization of HOI detection results and the cross-attention maps in one decoder layer on HICO-DET. Images in the first and second rows represent results for QPIC and QPIC+HQM, respectively. Best viewed in color.}
    \label{Figure:Vis_supp}
    \vspace{-1mm}
\end{figure*}

\begin{table*}[t]
\centering
\caption{Performance comparisons on HICO-DET.  ‘A’, ‘P’,
‘S’, and ‘L’ represent the appearance feature, human pose feature,
spatial feature, and language feature, respectively.}
 \vspace{-3mm}
\resizebox{0.8\textwidth}{!}{
\begin{tabular}{c|c|c|c|c|ccc|ccc  }
\hline
                &      &      &  & & \multicolumn{3}{c|}{DT Mode}  &\multicolumn{3}{c}{KO Mode}\\
& Methods  & Feature & Backbone &Detector & Full & Rare & Non-Rare  & Full & Rare & Non-Rare \\
\hline
\hline
\multirow{7}*{\rotatebox{90}{CNN-based}}
&InteractNet \cite{Gkioxari2017Detecting} &A & ResNet-50-FPN  &COCO &9.94  & 7.16     & 10.77     & -   & -   & - \\
&UnionDet \cite{kim2020uniondet} &A & ResNet-50-FPN  &HICO-DET &17.58  & 11.72  & 19.33  & 19.76  & 14.68  & 21.27 \\
& PD-Net \cite{zhong2020polysemy} &A+S+P+L & ResNet-152  &COCO &20.81 &15.90 &22.28 &24.78 &18.88 &26.54\\
& DJ-RN \cite{li2020detailed} &A+S+P+L & ResNet-50  &COCO &21.34 &18.53 &22.18 &23.69 &20.64 &24.60 \\
&PPDM \cite{liao2020ppdm} & A & Hourglass-104  &HICO-DET & 21.73 &13.78 &24.10  & 24.58   & 16.65   & 26.84\\
& GGNet \cite{zhong2021glance}     & A  & Hourglass-104  &HICO-DET &23.47 &16.48 &25.60 &27.36 &20.23 & 29.48 \\
& VCL \cite{hou2020visual} &A+S & ResNet-101  &HICO-DET &23.63 &17.21 &25.55 &25.98 &19.12 &28.03 \\

\hline
\hline
\multirow{9}*{\rotatebox{90}{DETR-based}}
&HOTR \cite{kim2021hotr} &A & ResNet-50  &COCO &23.46 &16.21 &25.62     & -   & -   & - \\
&HOI-Trans \cite{zou2021end}  &A  & ResNet-50 &HICO-DET & 23.46 & 16.91 & 25.41 & 26.15 & 19.24 & 28.22 \\
 &AS-Net \cite{kim2020uniondet} &A  & ResNet-50 &HICO-DET &28.87  & 24.25  & 30.25  & 31.74  & 27.07  & 33.14 \\
&QPIC  \cite{tamura2021qpic} &A  & ResNet-50  &HICO-DET &29.07 &21.85 &31.23  & 31.68   & 24.14   & 33.93 \\
&ConditionDETR \cite{ConditionDetr} &A  & ResNet-50  &HICO-DET&29.65 &22.64 &31.75 &32.11 &24.62  & 34.34   \\
&CDN-S \cite{zhang2021mining}   & A  & ResNet-50 &HICO-DET &31.44 & 27.39 & 32.64 &34.09 &29.63 &35.42 \\
 &HOTR \cite{kim2021hotr}  + \textbf{HQM}    & A  & ResNet-50  &COCO & \textbf{25.69} & \textbf{24.70} &  \textbf{25.98} & \textbf{28.24} & \textbf{27.35} & \textbf{28.51} \\
&QPIC \cite{tamura2021qpic} + \textbf{HQM}  & A & ResNet-50  &HICO-DET &\textbf{31.34} &\textbf{26.54} &\textbf{32.78}  & \textbf{34.04}  & \textbf{29.15}  & \textbf{35.50} \\
& CDN-S \cite{zhang2021mining}   + \textbf{HQM}  & A  & ResNet-50  &HICO-DET & \textbf{32.47} & \textbf{28.15} & \textbf{33.76}  & \textbf{35.17} & \textbf{30.73} & \textbf{36.50} \\
\hline
\end{tabular}}
\vspace{-3mm}
\label{tab:hico}
\end{table*}

\section{Performance Comparisons on HICO-DET}
\label{sec:hico}
We here present the complete comparisons between our method and state-of-the-arts on HICO-DET in Table \ref{tab:hico}.

\section{Qualitative Visualization Results}
\label{vis}
Fig. \ref{Figure:Vis_supp} provides more qualitative comparisons between
QPIC \cite{tamura2021qpic} and QPIC + HQM  in terms of cross-attention maps and HOI detection results on HICO-DET. We can observe that HQM enables QPIC to capture more discriminative image areas for HOI prediction.
\end{appendix}

\end{document}